\documentclass{article}

\PassOptionsToPackage{numbers, sort&compress}{natbib}
\usepackage[eandd, final]{neurips_2026}

\makeatletter
\def\@noticestring{}
\makeatother

\usepackage[utf8]{inputenc} 
\usepackage[T1]{fontenc}    
\usepackage{hyperref}       
\usepackage{xurl}           
\usepackage{booktabs}       
\usepackage{amsfonts}       
\usepackage{nicefrac}       
\usepackage{microtype}      
\usepackage[table]{xcolor}  
\usepackage[varqu,varl]{inconsolata} 
\usepackage{etoolbox}

\usepackage{amsmath,amssymb,amsthm}
\usepackage{tabularx}
\usepackage{multirow}

\usepackage{graphicx}
\usepackage{array}
\usepackage{tikz}
\usepackage{svg}
\svgsetup{inkscapepath=svg-inkscape/} 

\usepackage{float}
\usepackage{placeins}

\usepackage[most]{tcolorbox}
\usepackage{listings}
\usepackage{caption}
\usepackage{enumitem}

\usepackage{pifont}

\definecolor{RPPrimary}{HTML}{2F67B1}
\definecolor{RPDeep}{HTML}{234A78}
\definecolor{RPCyan}{HTML}{18A8D8}
\definecolor{RPCoral}{HTML}{E88A63}
\definecolor{RPBlueTint}{HTML}{EAF3FB}
\definecolor{RPCoralTint}{HTML}{FDF0E9}
\definecolor{RPInk}{HTML}{263445}
\definecolor{RPMuted}{HTML}{6B7280}
\definecolor{RPGrayTint}{HTML}{F4F7FA}

\definecolor{pbGreen}{HTML}{2E9D67}
\definecolor{pbRed}{HTML}{C85A5A}
\definecolor{pbPartial}{HTML}{5FA3E8}

\hypersetup{
  colorlinks=true,
  linkcolor=RPPrimary,
  citecolor=RPDeep,
  urlcolor=RPCyan,
  pdfborder={0 0 0}
}

\newcommand{\rphead}[1]{\noindent\textcolor{RPDeep}{\textbf{#1}}}
\newcommand{\rpstatichead}[1]{\noindent\textcolor{RPPrimary}{\textbf{#1}}}
\newcommand{\rpdynamichead}[1]{\noindent\textcolor{RPCoral}{\textbf{#1}}}
\newcommand{\rpwarninghead}[1]{\noindent\textcolor{RPCoral}{\textbf{#1}}}

\newtheoremstyle{mydefinition}
  {\topsep}   
  {\topsep}   
  {\upshape}  
  {}          
  {\color{RPDeep}\bfseries} 
  {.}         
  {.5em}      
  {\thmname{#1}\thmnumber{ #2}\thmnote{ (\textit{#3})}} 

\theoremstyle{mydefinition}

\definecolor{pbBlueLight}{HTML}{EAF3FB}
\definecolor{pbBlueMid}{HTML}{A9C9E8}
\definecolor{pbBlueDeep}{HTML}{4E86C3}
\definecolor{pbBlueDark}{HTML}{234A78}
\definecolor{pbBlueBorder}{HTML}{2F67B1}

\newcommand{\singleicon}{%
\tikz[baseline=-0.6ex]{
  \node[
    draw=pbBlueBorder,
    rounded corners=0.22ex,
    fill=pbBlueLight,
    minimum size=2.1ex,
    inner sep=0pt
  ] {};
}}

\newcommand{\orderedicon}{%
\tikz[baseline=-0.6ex]{
  \foreach \n/\x in {1/0,2/0.34,3/0.68,4/1.02}{
    \node[
      draw=pbBlueBorder,
      rounded corners=0.22ex,
      fill=pbBlueLight,
      minimum size=2.1ex,
      inner sep=0pt,
      text=pbBlueDark,
      font=\scriptsize\bfseries
    ] at (\x,0) {\n};
  }
}}

\newcommand{\shuffleicon}{%
\tikz[baseline=-0.6ex]{
  \node[
    draw=pbBlueBorder,
    rounded corners=0.22ex,
    fill=pbBlueLight,
    minimum size=2.1ex,
    inner sep=0pt
  ] at (0,0) {};
  \node[
    draw=pbBlueBorder,
    rounded corners=0.22ex,
    fill=pbBlueMid,
    minimum size=2.1ex,
    inner sep=0pt
  ] at (0.34,0) {};
  \node[
    draw=pbBlueBorder,
    rounded corners=0.22ex,
    fill=pbBlueDeep,
    minimum size=2.1ex,
    inner sep=0pt
  ] at (0.68,0) {};
}}

\newcommand{\compareicon}{%
\tikz[baseline=-0.6ex]{
  \node[
    draw=pbBlueBorder,
    rounded corners=0.22ex,
    fill=pbBlueLight,
    minimum size=2.1ex,
    inner sep=0pt
  ] at (0,0) {};
  \node[
    draw=pbBlueBorder,
    rounded corners=0.22ex,
    fill=pbBlueDeep,
    minimum size=2.1ex,
    inner sep=0pt
  ] at (0.34,0) {};
}}

\definecolor{pbBaseBest}{HTML}{A9C9E8}
\definecolor{pbBaseSecond}{HTML}{EAF3FB}
\definecolor{pbSFT}{HTML}{FDF0E9}

\newcommand{\basebest}[1]{\cellcolor{pbBaseBest}#1}
\newcommand{\basesecond}[1]{\cellcolor{pbBaseSecond}#1}
\newcommand{\sftcell}[1]{\cellcolor{pbSFT}#1}

\makeatletter
\newlength{\RPTitleTopGap}
\newlength{\RPTitleBottomGap}
\renewcommand{\@toptitlebar}{%
  {\color{RPPrimary}\rule{\linewidth}{3\p@}}%
  \par
  \vskip \RPTitleTopGap
  \vskip -\parskip%
}
\renewcommand{\@bottomtitlebar}{%
  \vskip \RPTitleBottomGap
  \vskip -\parskip
  {\color{RPPrimary}\hrule height 1.2\p@}%
  \vskip 0.08in%
}
\renewcommand{\section}{%
  \@startsection{section}{1}{\z@}%
                {-2.2ex \@plus -0.5ex \@minus -0.2ex}%
                {1.15ex \@plus 0.3ex \@minus 0.2ex}%
                {\color{RPDeep}\large\bfseries\raggedright}%
}
\renewcommand{\subsection}{%
  \@startsection{subsection}{2}{\z@}%
                {-1.9ex \@plus -0.5ex \@minus -0.2ex}%
                {0.7ex \@plus 0.2ex}%
                {\color{RPPrimary}\normalsize\bfseries\raggedright}%
}
\renewcommand{\subsubsection}{%
  \@startsection{subsubsection}{3}{\z@}%
                {-1.5ex \@plus -0.4ex \@minus -0.2ex}%
                {0.5ex \@plus 0.2ex}%
                {\color{RPDeep}\normalsize\bfseries\raggedright}%
}
\patchcmd{\@maketitle}
  {\LARGE\bf \@title\par}
  {\fontsize{14.5pt}{17.2pt}\selectfont\bfseries \@title\par}
  {}{}
\makeatother

\renewenvironment{abstract}{%
  \vskip 0.04in
  \begin{tcolorbox}[
    enhanced,
    breakable,
    colback=RPBlueTint!55!white,
    colframe=RPPrimary!38!white,
    colbacktitle=RPPrimary,
    coltitle=white,
    fonttitle=\bfseries,
    title={Abstract},
    attach boxed title to top left={xshift=9pt,yshift=-2.2mm},
    boxed title style={
      boxrule=0pt,
      arc=2pt,
      left=6pt,right=6pt,top=2pt,bottom=2pt
    },
    boxrule=0.55pt,
    arc=3pt,
    outer arc=3pt,
    left=9pt,right=9pt,top=10pt,bottom=7pt,
    before skip=5pt,after skip=9pt
  ]
  \small
}{%
  \end{tcolorbox}
}

\newtcolorbox{contributionbox}{
  enhanced,
  breakable,
  colback=RPGrayTint!72!white,
  colframe=RPPrimary!34!white,
  colbacktitle=RPDeep,
  coltitle=white,
  fonttitle=\bfseries,
  title={Contributions},
  attach boxed title to top left={xshift=8pt,yshift=-2.1mm},
  boxed title style={
    boxrule=0pt,
    arc=2pt,
    left=6pt,right=6pt,top=2pt,bottom=2pt
  },
  boxrule=0.5pt,
  arc=3pt,
  outer arc=3pt,
  left=8pt,right=8pt,top=9pt,bottom=5pt,
  before skip=7pt,after skip=7pt
}

\newtcolorbox{takeawaybox}{
  enhanced,
  colback=RPBlueTint!42!white,
  colframe=RPPrimary!34!white,
  colbacktitle=RPPrimary,
  coltitle=white,
  fonttitle=\bfseries,
  title={Diagnostic takeaway},
  attach boxed title to top left={xshift=8pt,yshift=-2.1mm},
  boxed title style={
    boxrule=0pt,
    arc=2pt,
    left=6pt,right=6pt,top=2pt,bottom=2pt
  },
  boxrule=0.5pt,
  arc=3pt,
  outer arc=3pt,
  left=8pt,right=8pt,top=9pt,bottom=6pt,
  before skip=7pt,after skip=7pt
}

\newtcolorbox{limitationbox}{
  enhanced,
  breakable,
  colback=RPCoralTint!45!white,
  colframe=RPCoral!38!white,
  boxrule=0.5pt,
  arc=3pt,
  outer arc=3pt,
  left=8pt,right=8pt,top=6pt,bottom=6pt,
  before skip=5pt,after skip=7pt
}

\title{\texorpdfstring{{\textsc{Robo2Process}}: Benchmarking and Learning Process-Aware\\Understanding in Vision-Language Robotic Manipulation}{Robo2Process: Benchmarking and Learning Process-Aware Understanding in Vision-Language Robotic Manipulation}}

\author{
{\color{RPInk}
\textbf{Dayu Xia}$^{2,*}$ \quad
\textbf{Yue Shi}$^{1,*,\dagger}$ \quad
\textbf{Yao Mu}$^{1,3,\dagger}$ \quad
\textbf{Huiting Ji}$^{5}$ \quad
\textbf{Chaofan Ma}$^{3}$} \\
{\color{RPInk}
\textbf{Yingjie Zhou}$^{3}$ \quad
\textbf{Jiezhang Cao}$^{3}$ \quad
\textbf{Hua Chen}$^{4}$ \quad
\textbf{Yang Liu}$^{2}$ \quad
\textbf{Guangtao Zhai}$^{1,3,\dagger}$}
\\[0.35em]
{\normalfont\color{RPMuted}
$^{1}$Shanghai AI Laboratory
\quad
$^{2}$Tsinghua University
\quad
$^{3}$Shanghai Jiao Tong University}
\\
{\normalfont\color{RPMuted}
$^{4}$LimX Dynamics
\quad
$^{5}$China University of Mining and Technology}
\\[0.18em]
{\normalfont\small\color{RPDeep}
$^{*}$Equal contribution \quad
$^{\dagger}$Corresponding author} \\[0.15em]
}

\begin{document}

\raggedbottom

\maketitle
\vspace{-0.42in}

\begin{figure}[H]
  \centering
  \IfFileExists{Fig1.pdf}{%
    \includegraphics[width=0.98\linewidth]{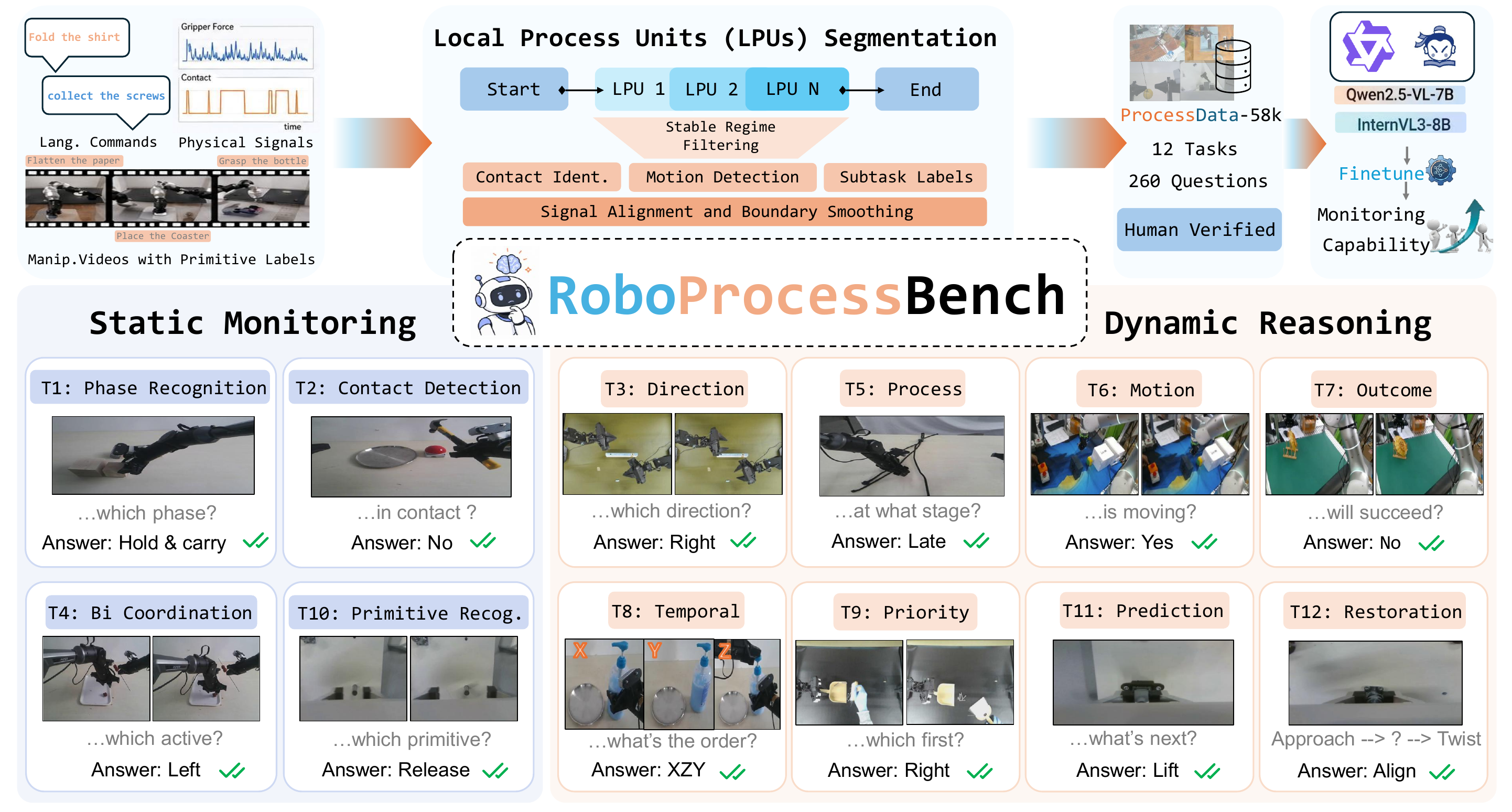}%
  }{%
    \fbox{\parbox[c][0.16\textheight][c]{0.94\linewidth}{\centering\small Figure asset \texttt{Fig1.pdf} is not present in the current workspace.}}%
  }
  \captionsetup{font=footnotesize,skip=3pt}
  \caption{\textbf{Overview of Robo2Process and RoboProcessBench.} Robo2Process converts source-native execution traces into local process units and structured QA supervision. RoboProcessBench instantiates 12 question families across static monitoring and dynamic reasoning.}
  \label{fig:RoboProcessBench_overview}
  \vspace{-0.35cm}
\end{figure}

\begin{abstract}
Vision-language models (VLMs) used as robotic critics, reward generators, and failure detectors must understand how manipulation unfolds, not only whether it succeeds. Yet fine-grained process supervision is not readily available in a unified form, and existing evaluations do not systematically diagnose these abilities. We introduce Robo2Process, a trace-grounded framework for scalable construction of process-aware supervision from heterogeneous robotic execution data. It aligns visual observations with source-native signals and annotations, extracts local process units (LPUs), and generates structured QA supervision through task-specific sampling, deterministic label construction, and ambiguity filtering. We instantiate this framework as RoboProcessBench, a diagnostic benchmark with 12 question families spanning \emph{static monitoring} and \emph{dynamic reasoning}, and ProcessData, a corpus of 57,892 QA pairs across four data sources and 260 manipulation tasks, split into ProcessData-SFT and ProcessData-Eval. Evaluation of a broad suite of VLMs reveals uneven process understanding. Supervision from ProcessData-SFT improves \textit{Qwen2.5-VL-7B} and \textit{InternVL-3-8B} on local state, motion, progress, and primitive-aware cues. External evaluation on RoboRewardBench shows Robo2Process enables out-of-distribution process-aware reasoning abilities. Overall, Robo2Process provides a source-aware pipeline, diagnostic benchmark, and learnable substrate for process-aware understanding in robotic manipulation.

\smallskip
\noindent\textcolor{RPDeep}{\textbf{Project page}}\enspace
\href{https://processbench-2026.github.io/RoboProcessBench-Web/}{\texttt{processbench-2026.github.io}}

\end{abstract}

\section{Introduction}
Vision-language models (VLMs) are increasingly used as visual critics, online reward generators, and failure-reasoning modules in robotic manipulation \citep{zhao2024vlmpc, zhai2025vlac, ji2026prmjudge, wu2026large, duan2024aha, tan2025robodopamine}. These roles require more than recognizing objects or predicting a final success label: a useful process judge must assess whether contact has been established, motion is active, a local primitive is progressing, temporal evidence is consistent, and the current state supports the next step.

Recent work studies subgoal-level evaluation, process reward modeling, progress-aware critics, and failure reasoning \citep{elmallah2025stepeval, ji2026prmjudge, wu2026large, duan2024aha, zhai2025vlac}. These systems demonstrate the value of intermediate signals, but generally focus on deploying a judge for policy auditing, reward generation, or failure analysis. A complementary gap is the lack of a reusable way to derive process-understanding supervision from heterogeneous robotic data, where relevant evidence may be distributed across visual observations, force or torque traces, gripper state, timestamps, success records, and primitive annotations.

We address these gaps with \textbf{Robo2Process}, a source-aware pipeline for constructing process supervision from native execution signals and annotations. Robo2Process aligns observations with source-native evidence, extracts \emph{local process units} (LPUs), and applies task-specific sampling, deterministic label construction, and ambiguity guards. This enables batch generation and reuse across eligible episodes without manually labeling every QA pair, while still requiring source-specific setup and reliable native anchors.

We instantiate the pipeline as \textbf{RoboProcessBench}, a 12-family diagnostic benchmark, and \textbf{ProcessData}, a corpus of 57,892 QA pairs from four robotic sources and 260 manipulation tasks~\citep{aist2025bimanual,wang2026gm100,chen2024rh20tp,sliwowski2025reassemble}. Episode- and recording-isolated ProcessData-SFT and Eval splits support post-training and held-out diagnosis. Experiments reveal uneven VLM capabilities and show that post-training improves local state, motion, progress, and primitive-aware performance on two backbones, while temporal reasoning remains to be tackled. We additionally report external reward-prediction associations and mixed-training results on RoboRewardBench.

\begin{contributionbox}
\begin{enumerate}[
  leftmargin=1.75em,
  label=\textcolor{RPPrimary}{\bfseries\arabic*.},
  itemsep=2pt,
  topsep=1pt,
  parsep=0pt
]
\item \textbf{Robo2Process}, a source-aware pipeline for batch construction of process supervision from native execution signals and annotations, using LPUs, task-specific sampling, deterministic label builders, and ambiguity guards.
\item \textbf{RoboProcessBench}, a diagnostic benchmark spanning 12 process-understanding families across static monitoring and dynamic reasoning.
\item \textbf{ProcessData}, a four-source instantiation with 57,892 QA pairs, 260 manipulation tasks, and isolated SFT and Eval splits.
\item Multi-level validation through human audit, VLM diagnosis, post-training on two backbones, and external reward evaluation, characterizing demonstrated benefits and limitations.
\end{enumerate}
\end{contributionbox}

\section{Related Work}

\rphead{Robot Data and Benchmarks.}
Large-scale robot datasets and embodied benchmarks have substantially advanced manipulation research by providing diverse trajectories, tasks, embodiments, and environments \citep{brohan2023openx,khazatsky2024droid,bu2025agibotworld,wang2026gm100,wu2024robomind,chen2024rh20tp,sliwowski2025reassemble,aist2025bimanual}. Existing benchmarks evaluate broad embodied capabilities such as visual-spatial reasoning, planning, language-conditioned manipulation, low-level action reasoning, and trajectory-conditioned question answering \citep{liu2023libero,haresh2024clevrskills,zhang2024vlabench,yang2024think,yang2025embodiedbench,zhao2025manipbench,chen2025robo2vlm,baai2025robobrain2,luo2025vebrain}. These resources are essential for policy learning and embodied evaluation, but they usually treat manipulation process structure as auxiliary information rather than the central evaluation target. In contrast, RoboProcessBench converts physically grounded execution traces into a VLM-side benchmark for process-aware understanding, covering contact, phase, motion, local progress, temporal relation, outcome, and primitive-level cues.

\rphead{Process and Reward Signals.}
A growing line of work argues that final success rates are insufficient for understanding how a robot policy executes a task. Step-level and subgoal-level evaluation exposes partial competence that binary outcomes hide \citep{elmallah2025stepeval}, while execution monitoring and precondition--effect modeling study whether intermediate action states are valid or anomalous \citep{sliwowski2025conditionnet}. Recent process reward and progress-aware methods further use intermediate execution signals for policy evaluation, reward generation, or control \citep{zhai2025vlac,wu2026large,tan2025robodopamine,yan2026progressvla,dai2026spr}. Closely related resources also provide intermediate representations, primitive-level annotations, or hierarchical manipulation structure for fine-grained reasoning \citep{li2026robointer,chen2024rh20tp,chen2025robohiman,wang2025roboeval,liu2026primor1}. These works motivate the importance of process signals, but their primary focus is typically reward modeling, monitoring, control, or representation learning. RoboProcessBench instead uses process signals as diagnostic and trainable supervision for evaluating whether VLMs understand the physical and temporal evolution of manipulation.

\rphead{VLMs as Robotic Judges.}
VLMs are increasingly used as active feedback modules in robotic systems. VLMPC uses VLM-based evaluation for action selection and future-state assessment \citep{zhao2024vlmpc}; VLAC learns a vision-language-action critic that predicts progress deltas and completion signals \citep{zhai2025vlac}; Large Reward Models adapt foundation VLMs into online reward generators \citep{wu2026large}; and AHA detects and reasons over manipulation failures to support downstream robotic frameworks \citep{duan2024aha}. Most directly related, PRM-as-a-Judge formulates dense robotic auditing with process reward models, introduces OPD metrics for outcome, process, and diagnosis-level evaluation, and uses RoboPulse to test fine-grained progress discrimination \citep{ji2026prmjudge}. These works show that VLMs and PRMs can serve as critics, reward models, progress monitors, and failure detectors. However, they usually instantiate or evaluate a judge within a downstream pipeline. RoboProcessBench addresses the complementary prerequisite: it decomposes the VLM-side process understanding needed by such judges into explicit benchmark tasks and trainable supervision signals.

\section{Robo2Process: Trace-Grounded Process Supervision}
\label{sec:RoboProcessBench}

Robo2Process converts heterogeneous robotic execution data into process-aware supervision through a shared episode--LPU--QA interface. Figure~\ref{fig:RoboProcessBench_overview} summarizes the construction pipeline and its RoboProcessBench task instantiation. Source-native traces ground labels, visual observations form the model inputs, and the resulting QA pairs support diagnostic evaluation and supervised adaptation across state, contact, motion, progress, temporal, outcome, and primitive-level cues.

\subsection{Overview and Evaluation Target}

Given a manipulation episode, RoboProcessBench does not focus only on final task success. Instead, it asks whether a VLM can recognize and reason about the intermediate process structure of the execution. Formally, let a manipulation episode be
\begin{equation}
e = \{(I_t, z_t)\}_{t=1}^{T},
\end{equation}
where \(I_t\) denotes the visual observation at time \(t\), and \(z_t\) denotes source-native episode execution information available at the same instant, such as task metadata, segment annotations, force and torque signals, gripper state, joint motion, velocity statistics, timestamps, or success and failure records.

For a task family \(m\), let \(z_t^{\star}\) denote the task-relevant latent process state and let
\begin{equation}
y = \phi_m(z_t^{\star})
\end{equation}
be the corresponding benchmark target, such as phase, contact status, motion state, local progress, temporal relation, or primitive-level state. A model \(f_\theta\) is evaluated by predicting
\begin{equation}
\hat{y} = f_\theta(x,c),
\end{equation}
where \(x\) is the visual input and \(c\) is optional task context. The target \(y\) is therefore a structured prediction about the ongoing manipulation process rather than a generic scene description. Overall, RoboProcessBench evaluates the perceptual and temporal process variables that a PRM-style judge must implicitly estimate. For example, progress rewards require sensitivity to local subprogress, regression, and temporal direction; failure detectors require contact, phase, and outcome cues; and primitive-level critics require knowledge of current and next local actions. Thus, RoboProcessBench treats process-aware QA not as an end in itself, but as a structured supervision interface for training VLM-based process evaluators.

\rphead{Intended use.}
RoboProcessBench is intended for diagnosing and training VLM-side process understanding in robotic manipulation. It should be used to compare models across process-aware subskills and to study whether structured process supervision improves held-out process QA performance. 

\subsection{Anchored Local Process Units}

A key design choice of RoboProcessBench is that long manipulation episodes are decomposed into \emph{local process units} (LPUs):
\begin{equation}
\mathcal{U}(e)=\{u_j\}_{j=1}^{N_e}, \qquad
u_j=[t_j^s,t_j^e], \quad t_j^s<t_j^e.
\end{equation}
Each LPU is a temporally localized decision unit corresponding to a coherent manipulation subprogress. In contact-rich sources, LPUs are contact-anchored and derived from contact onset, contact persistence, release, or stable interaction regimes. In sources with explicit low-level action annotations, LPUs can additionally be primitive-anchored and aligned with native primitive intervals. This abstraction converts long manipulation episodes into auditable process-aware units, allowing the benchmark to evaluate local state, subprogress, and action-chain reasoning rather than only global episode completion. A benchmark item for task family \(m\) is defined as
\begin{equation}
x_i^{(m)} = \left(g_i,\mathbf{o}_i,q_i^{(m)},y_i^{(m)}\right),
\end{equation}
where \(g_i\) is the task context, \(\mathbf{o}_i\) is the extracted visual input, \(q_i^{(m)}\) is the task-specific question, and \(y_i^{(m)}\) is the ground-truth answer. The full benchmark is
\begin{equation}
\mathcal{B} = \bigcup_{m=1}^{12} \mathcal{B}_m,
\qquad
\mathcal{B}_m = \{x_i^{(m)}\}_{i=1}^{N_m}.
\end{equation}

\subsection{Question Taxonomy}

RoboProcessBench organizes 12 question families into two complementary categories, as summarized in Table~\ref{tab:RoboProcessBench_taxonomy}. \emph{Static Monitoring} evaluates whether a VLM can identify the current local process state, while \emph{Dynamic Reasoning} evaluates whether it can infer motion, progress, temporal relation, outcome, or primitive-level continuation.

\rpstatichead{Static process monitoring.}
Static monitoring families ask the model to identify the current local process state from a single frame or a short ordered clip. These include phase recognition (\textbf{\texttt{T1}}), contact detection (\textbf{\texttt{T2}}), bimanual coordination state recognition (\textbf{\texttt{T4}}), and current primitive recognition (\textbf{\texttt{T10}}). Although these tasks may appear visually local, they are process-aware rather than object-centric: the VLM must infer which manipulation-relevant state has already been reached.

\rpdynamichead{Dynamic process reasoning.}
Dynamic reasoning families require temporal, predictive, progress-centric, or chain-level inference. These include motion direction prediction (\textbf{\texttt{T3}}), primitive-local progress recognition (\textbf{\texttt{T5}}), motion state recognition (\textbf{\texttt{T6}}), operation outcome prediction (\textbf{\texttt{T7}}), temporal ordering (\textbf{\texttt{T8}}), temporal priority prediction (\textbf{\texttt{T9}}), next primitive prediction (\textbf{\texttt{T11}}), and primitive chain restoration (\textbf{\texttt{T12}}). Together, these families test whether a VLM can infer how the current state is evolving and how the local process should continue.

We treat \textbf{\texttt{T10}}--\textbf{\texttt{T12}} as primitive-aware extensions in the current release, because they rely on explicit native primitive annotations. The remaining families are cross-dataset process families constructed from source-native physical, temporal, or outcome signals.

\vspace{0.5cm}

\begin{table*}[t]
\caption{\textbf{RoboProcessBench task taxonomy.} The 12 tasks cover current-state monitoring, temporal reasoning, and primitive-aware extensions. Under \textit{Input Frame}, different colors indicate shuffled frames, while ordered frames are annotated with numbers.}
\label{tab:RoboProcessBench_taxonomy}
\centering
\small
\setlength{\tabcolsep}{5pt}
\renewcommand{\arraystretch}{1.10}
\begin{tabularx}{\textwidth}{c l X c}
\toprule
\textbf{ID} & \textbf{Task} & \textbf{Question} & \textbf{Input Frame} \\
\midrule
\rowcolor{RPBlueTint!45!white}
T1 & Phase Recognition & What coarse process phase is the manipulation currently in? & \singleicon \\
\rowcolor{RPBlueTint!45!white}
T2 & Contact Detection & Has effective task-relevant contact been established? & \singleicon \\
\rowcolor{RPBlueTint!45!white}
T4 & Bimanual Coordination State & How are the two arms currently coordinating? & \orderedicon \\
\rowcolor{RPBlueTint!45!white}
T10$^\dagger$ & Current Primitive Recognition & What low-level primitive is being executed now? & \orderedicon \\
\midrule
\rowcolor{RPCoralTint!48!white}
T3 & Motion Direction Prediction & What is the dominant motion direction? & \orderedicon \\
\rowcolor{RPCoralTint!48!white}
T5 & Primitive-local Progress & Is the current local step early, middle, or late? & \orderedicon \\
\rowcolor{RPCoralTint!48!white}
T6 & Motion State Recognition & Is the manipulator actively moving or effectively stationary? & \orderedicon \\
\rowcolor{RPCoralTint!48!white}
T7 & Operation Outcome Prediction & Will the ongoing attempt eventually succeed? & \orderedicon \\
\rowcolor{RPCoralTint!48!white}
T8 & Temporal Ordering & What is the correct chronological order of the three frames? & \shuffleicon \\
\rowcolor{RPCoralTint!48!white}
T9 & Temporal Priority Prediction & Which of the two frames happened earlier? & \compareicon \\
\rowcolor{RPCoralTint!48!white}
T11$^\dagger$ & Next Primitive Prediction & What low-level primitive should happen next? & \orderedicon \\
\rowcolor{RPCoralTint!48!white}
T12$^\dagger$ & Primitive Chain Restoration & Which primitive best fills the masked slot in the local action chain? & \orderedicon \\
\bottomrule
\end{tabularx}

\vspace{2pt}
\raggedright\footnotesize
$^\dagger$ Primitive-aware extension tasks grounded in explicit \textbf{REASSEMBLE} annotations.
\end{table*}
\vspace{-0.5cm}

\subsection{Benchmark Construction}

Top of Figure~\ref{fig:RoboProcessBench_overview} illustrates the construction pipeline. Ground-truth construction follows three principles: source-native supervision is preferred over free-form manual annotation; labels are constructed only after identifying task-relevant LPUs; and human intervention is used for calibration and audit rather than exhaustive relabeling. For a task family \(m\), let \(\Pi_m\) denote the task-specific visual extractor and \(\Gamma_m\) the corresponding label constructor. Each item is instantiated as
\begin{equation}
\mathbf{o}_i = \Pi_m(e_i,u_i), \qquad
y_i^{(m)} = \Gamma_m(e_i,u_i,g_i),
\end{equation}
where \(u_i\) is the associated local process unit. The extractor \(\Pi_m\) determines whether the visual input is a single frame, an ordered clip, a shuffled frame set, or a pairwise comparison. The constructor \(\Gamma_m\) maps source-native signals and metadata to a discrete answer. RoboProcessBench does not impose a single universal frame rule. Instead, each task family uses the minimal temporal context required by its decision scale. \textbf{\texttt{T1}} and \textbf{\texttt{T2}} use single frames; \textbf{\texttt{T3}}, \textbf{\texttt{T4}}, \textbf{\texttt{T5}}, \textbf{\texttt{T6}}, \textbf{\texttt{T7}}, \textbf{\texttt{T10}}, \textbf{\texttt{T11}}, and \textbf{\texttt{T12}} use short ordered clips; \textbf{\texttt{T8}} uses three shuffled frames; and \textbf{\texttt{T9}} uses a two-frame comparison. This design avoids making process reasoning artificially easy by overexposing context or artificially hard by removing all temporal evidence.

At a high level, the 12 task families follow four construction regimes. \textbf{\texttt{T1}}, \textbf{\texttt{T2}}, \textbf{\texttt{T4}}, and \textbf{\texttt{T6}} are built from stable state, contact, or motion regimes; \textbf{\texttt{T3}}, \textbf{\texttt{T8}}, and \textbf{\texttt{T9}} are built from motion and timestamp structure; \textbf{\texttt{T5}} and \textbf{\texttt{T7}} are built from local subprogress and eventual outcome signals; and \textbf{\texttt{T10}}--\textbf{\texttt{T12}} are instantiated from explicit primitive annotations when available.

\begin{figure*}[t]
    \centering  
    \IfFileExists{Fig2.pdf}{%
      \includegraphics[width=0.98\linewidth]{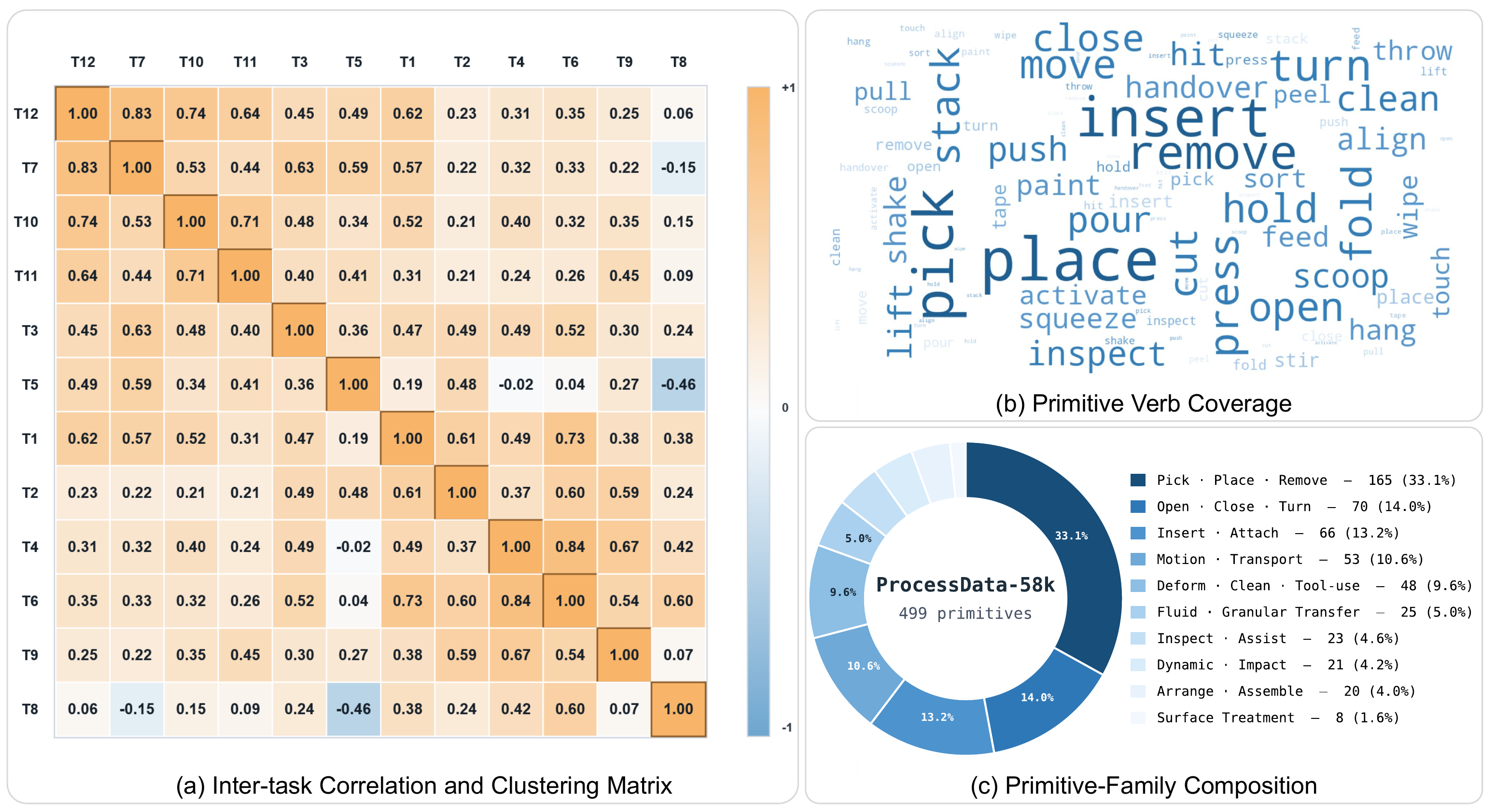}%
    }{%
      \fbox{\parbox[c][0.14\textheight][c]{0.94\linewidth}{\centering\small Figure asset \texttt{Fig2.pdf} is not present in the current workspace.}}%
    }
    \caption{\textbf{Inter-task structure and primitive coverage in RoboProcessBench.} (a) Spearman correlation and hierarchical clustering of chance-normalized family accuracies for the original 16-model result set (14 zero-shot baselines and two post-trained models). (b) Primitive verb coverage. (c) Primitive-family composition in ProcessData.}
    \label{fig:RoboProcessBench_profile_coverage}
    \vspace{0.3cm}
\end{figure*}

\subsection{Manual audit}
Manual effort is intentionally limited and high-leverage. Robo2Process does not rely on dense frame-wise manual labeling; human input includes source-specific setup, task-level primitive chains, object slots, and reliability audit. For each task family, 50 items sampled from the full ProcessData corpus (SFT and Eval) were independently answered by two annotators. Average accuracy exceeds 90\% across task families. This is a corpus-level answerability audit, not an Eval-only human baseline.

\section{ProcessData}
\label{sec:RoboProcessBench_dataset}

ProcessData is the data substrate of RoboProcessBench. It aggregates four complementary robotic manipulation sources~\citep{wang2026gm100,chen2024rh20tp,sliwowski2025reassemble,aist2025bimanual}, yielding 57,892 process-aware QA items across 260 manipulation tasks. Rather than treating all sources as a homogeneous pool, we use each dataset for the type of process supervision it supports best: \texttt{GM-100} provides long-tail goal-conditioned manipulation tasks, \texttt{RH20T} contributes contact-rich multimodal traces, \texttt{REASSEMBLE} provides explicit primitive-level action chains, and \texttt{AIST-Bimanual} supplies bimanual motion and coordination patterns. ProcessData is split into 48,841 SFT and 9,051 Eval items (an 85/15 target ratio) with strict episode- or recording-group isolation (seed 42).

This source design is essential for process-aware evaluation. RoboProcessBench requires labels tied to local execution structure, such as contact, motion state, local progress, temporal order, outcome, and primitive transitions, which may be non-uniform across sources. We therefore apply a unified trace-driven construction pipeline over heterogeneous native signals and annotations: visual observations are aligned with execution traces, contact- or primitive-anchored local process units are extracted, and task-specific QA items are instantiated. 

Table~\ref{tab:processdata_source_coverage} summarizes how the four sources jointly support the 12-task taxonomy. The goal of this aggregation is to combine complementary forms of supervision into a unified benchmark-ready corpus for process-aware manipulation understanding. Complementing the source-level view in Table~\ref{tab:processdata_source_coverage}, the right panels of Figure~\ref{fig:RoboProcessBench_profile_coverage} summarize the primitive-level semantic coverage of ProcessData, including both frequent primitive verbs and their aggregated family-level composition. This visualization shows that ProcessData spans a diverse set of reusable local process units.

\definecolor{pbCover}{HTML}{A9C9E8}
\definecolor{pbCoverLight}{HTML}{EAF3FB}
\definecolor{pbEmpty}{HTML}{F4F7FA}
\definecolor{pbText}{HTML}{234A78}

\newcommand{\srcyes}{\cellcolor{pbCover}\textcolor{pbText}{\scriptsize\bfseries \checkmark}}
\newcommand{\srcno}{\cellcolor{pbEmpty}{\scriptsize --}}

\begin{table*}[t]
\centering
\small
\setlength{\tabcolsep}{4pt}
\renewcommand{\arraystretch}{1.08}
\caption{\textbf{Source contribution to ProcessData.}
Each source is used only for the RoboProcessBench task families supported by its native signals or annotations, yielding complementary coverage across the 12-task taxonomy.}
\label{tab:processdata_source_coverage}
\resizebox{\textwidth}{!}{
\begin{tabular}{l l cccc cccccccc}
\toprule
\textbf{Source} & \textbf{Main Role} 
& \multicolumn{4}{c}{\textbf{Static Monitoring}} 
& \multicolumn{8}{c}{\textbf{Dynamic Reasoning}} \\
\cmidrule(lr){3-6} \cmidrule(lr){7-14}
& & \textbf{T1} & \textbf{T2} & \textbf{T4} & \textbf{T10} & \textbf{T3} & \textbf{T5} & \textbf{T6} & \textbf{T7} & \textbf{T8} & \textbf{T9} & \textbf{T11} & \textbf{T12} \\
\midrule
GM-100 & Long-tail goal-conditioned manipulation 
& \srcyes & \srcyes & \srcyes & \srcno 
& \srcyes & \srcyes & \srcyes & \srcno & \srcyes & \srcyes & \srcno & \srcno \\

RH20T & Contact-rich multimodal traces 
& \srcyes & \srcyes & \srcno & \srcno 
& \srcyes & \srcyes & \srcyes & \srcyes & \srcyes & \srcyes & \srcno & \srcno \\

REASSEMBLE & Primitive-level action chains 
& \srcyes & \srcyes & \srcno & \srcyes 
& \srcno & \srcyes & \srcyes & \srcyes & \srcyes & \srcyes & \srcyes & \srcyes \\

AIST-Bimanual & Bimanual motion and coordination 
& \srcno & \srcno & \srcyes & \srcno 
& \srcyes & \srcno & \srcyes & \srcno & \srcyes & \srcyes & \srcno & \srcno \\
\midrule
\textbf{ProcessData} & Unified process-aware QA corpus 
& \srcyes & \srcyes & \srcyes & \srcyes 
& \srcyes & \srcyes & \srcyes & \srcyes & \srcyes & \srcyes & \srcyes & \srcyes \\
\bottomrule
\end{tabular}
}
\end{table*}

\section{Experiments}
\label{sec:experiments}

We evaluate RoboProcessBench with two goals: diagnosing fine-grained process understanding on ProcessData-Eval and testing whether ProcessData-SFT provides learnable supervision for held-out VLM adaptation. The main results table reports 18 models (16 baselines without ProcessData-SFT adaptation and two ProcessData-SFT models). We also analyze the correlation structure of the 12 task families to assess and validate the task family splitting.

\subsection{Experimental Protocol}
\label{subsec:exp_protocol}

We evaluate a broad set of open-source and closed-source VLMs. The open-source baselines include QwenVL-family models~\citep{bai2025qwen25vl,qwen2025qwen3vl}, InternVL-family models~\citep{zhu2025internvl3,wang2025internvl35}, RoboBrain~\citep{baai2025robobrain2}, MiMo-Embodied~\citep{xiaomi2025mimoembodied}, Embodied-R1.5~\citep{yuan2026embodiedr15}, and GLM-4.6V~\citep{zai2025glm46v}. The closed-source baselines include Gemini~\citep{google2026gemini3flash}, GPT variants~\citep{openai2024gpt4o,openai2026gpt54mini}, and Claude variants~\citep{anthropic2025claudehaiku45,anthropic2026claudesonnet46}. We also report two post-trained VLMs, \textit{ProcessData-SFT-Qwen} and \textit{ProcessData-SFT-Intern}, obtained by supervised fine-tuning on ProcessData-SFT.

All tasks are formulated as multiple-choice visual question answering, and results are reported as accuracy on ProcessData-Eval. RoboProcessBench uses strict episode and recording isolation between ProcessData-SFT and ProcessData-Eval to reduce leakage from near-duplicate trajectories and to ensure that evaluation reflects generalization across execution instances rather than memorization of specific recordings. For post-training, we fine-tune \textit{Qwen2.5-VL-7B} and \textit{InternVL-3-8B} with LoRA using rank 8 and alpha 16. The language model is fine-tuned while the vision encoder is kept frozen. Fine-tuning uses 4 NVIDIA H200 GPUs, and all evaluations are run on 1 NVIDIA H200 GPU. We use temperature 0.01 and a maximum output length of 512 tokens for all evaluated models.

\subsection{Diagnostic Results on ProcessData-Eval}
\label{subsec:main_results}

Table~\ref{tab:main_results} reports the full 12-task evaluation for 16 baselines without ProcessData-SFT adaptation and two post-trained models. Figure~\ref{fig:RoboProcessBench_profile_coverage}(a) retains the earlier 16-model correlation analysis (14 zero-shot baselines and two post-trained models): related probes cluster strongly (\textbf{\texttt{T4}} and \textbf{\texttt{T6}}, $\rho=0.844$; \textbf{\texttt{T10}}--\textbf{\texttt{T12}}, $\rho=0.635$--$0.737$), whereas \textbf{\texttt{T8}} is nearly uncorrelated with \textbf{\texttt{T9}} and \textbf{\texttt{T12}} ($\rho\approx0.06$) and negatively correlated with \textbf{\texttt{T5}} ($\rho=-0.463$). The two newly added baselines, MiMo-Embodied and Embodied-R1.5, are not included in this existing correlation figure. Thus, related families align when expected, while temporal ordering, temporal priority, primitive restoration, and progress estimation expose separable failure modes in that 16-model analysis.

\rphead{Zero-shot strengths are fragmented across task families.}
No zero-shot model dominates all task families. The strongest baseline differs across temporal ordering (\textbf{\texttt{T8}}), temporal-priority prediction (\textbf{\texttt{T9}}), next primitive prediction (\textbf{\texttt{T11}}), outcome prediction (\textbf{\texttt{T7}}), and motion direction prediction (\textbf{\texttt{T3}}). This fragmented pattern indicates that RoboProcessBench reveals model-specific process-understanding profiles: different VLM families capture different subsets of local state, motion, temporal, and primitive-level cues.

\rphead{Local process-state grounding is easier than local progress estimation.}
Tasks grounded in relatively visible local states, such as contact detection (\textbf{\texttt{T2}}), motion-state recognition (\textbf{\texttt{T6}}), and primitive-chain restoration (\textbf{\texttt{T12}}), are more accessible to zero-shot VLMs than tasks requiring explicit estimation of process evolution. In contrast, primitive-local progress recognition (\textbf{\texttt{T5}}) remains difficult: the strongest zero-shot result reaches only \(34.4\%\), close to the \(33.3\%\) random baseline. This gap is central to the benchmark objective. \textbf{\texttt{T5}} asks whether a model can judge how far a local manipulation step has advanced, rather than merely identify what state is currently visible.

\rpwarninghead{Temporal reconstruction remains the hardest regime.}
Temporal ordering (\textbf{\texttt{T8}}) and temporal-priority prediction (\textbf{\texttt{T9}}) remain close to chance for most zero-shot models. The best zero-shot score on \textbf{\texttt{T8}} is only \(22.3\%\), and \textbf{\texttt{T9}} remains near the \(50\%\) random baseline. These results indicate that short visual context does not automatically yield reliable temporal reasoning in robotic manipulation. The weakness is therefore not just a low-score artifact. It identifies a key missing capability for VLMs expected to serve as progress monitors or process judges. 

\begin{table*}[t]
\centering
\footnotesize 
\setlength{\tabcolsep}{5pt} 
\caption{\textbf{Comprehensive evaluation on RoboProcessBench.} Accuracy (\%) on ProcessData-Eval for 16 baselines and two ProcessData-SFT models. Blue and light-blue cells mark the best and second-best results within each task family. Post-trained model scores are highlighted in light orange.}
\label{tab:main_results}
\begin{tabular*}{\textwidth}{@{\extracolsep{\fill}}l*{12}{c}@{}}
\toprule
\textbf{Model} & \multicolumn{4}{c}{\textcolor{RPDeep}{\textbf{Static Monitoring}}} & \multicolumn{8}{c}{\textcolor{RPCoral}{\textbf{Dynamic Reasoning}}} \\
\cmidrule(r){2-5} \cmidrule(lr){6-13}
& \textbf{T1} & \textbf{T2} & \textbf{T4} & \textbf{T10} & \textbf{T3} & \textbf{T5} & \textbf{T6} & \textbf{T7} & \textbf{T8} & \textbf{T9} & \textbf{T11} & \textbf{T12} \\
\midrule
\multicolumn{13}{@{}l}{\textit{Open-Source VLMs}} \\
Qwen2.5-VL-7B       & 26.6 & 41.9 & 36.4 & 33.1 & 35.5 & 32.2 & 58.0 & 54.6 & 17.9 & 50.9 & 33.1 & 80.4  \\
Qwen3-VL-8B         & \basesecond{34.1} & 42.3 & 37.5 & 32.9 & 30.1 & 32.7 & 64.7 & 60.7 & 15.9 & 52.2 & 59.0 & 84.8 \\
Qwen3-VL-32B        & 28.4 & \basesecond{53.5} & \basebest{41.9} & 28.1 & \basesecond{50.7} & 32.4 & \basesecond{69.2} & 50.6 & 19.7 & \basebest{53.4} & 47.0 & 76.1 \\
InternVL-3-8B       & 31.3 & 45.6 & 22.9 & 31.2 & 31.1 & 34.1 & 54.0 & \basesecond{61.9} & 17.3 & 48.7 & 46.5 & \basebest{91.3} \\
InternVL-3.5-8B     & \basebest{37.4} & 44.3 & 26.5 & \basesecond{36.8} & 27.7 & \basesecond{34.3} & 53.8 & \basesecond{61.9} & 17.5 & 50.7 & 53.5 & 80.4 \\
InternVL-3-38B      & 24.6 & 44.0 & 25.1 & 26.2 & 33.2 & \basebest{34.4} & 58.1 & 55.9 & 17.1 & 50.6 & \basesecond{63.5} & 82.6 \\
InternVL-3.5-38B    & 22.4 & 46.6 & 29.5 & 32.6 & 33.1 & 34.1 & 53.1 & 49.6 & 15.3 & 51.0 & 61.5 & 80.4 \\
RoboBrain-2.0-7B    & 29.8 & 44.0 & 26.6 & 32.6 & 44.4 & 33.8 & 51.2 & \basebest{62.5} & 15.8 & 49.4 & 32.6 & \basebest{91.3} \\
MiMo-Embodied-7B    & 28.4 & 43.1 & 38.8 & 27.9 & 32.1 & 30.5 & 67.2 & 59.1 & 11.9 & 48.5 & 52.5 & 78.3 \\
Embodied-R1.5-8B-SFT & 33.0 & \basebest{54.8} & 37.9 & 22.0 & 36.5 & 28.5 & \basebest{72.0} & 47.7 & 15.7 & 50.0 & 48.0 & \basebest{91.3} \\
GLM-4.6V             & 21.1 & 37.0 & 26.0 & 34.0 & 42.7 & 24.0 & 48.9 & 53.0 & 16.5 & 49.3 & 60.0 & 76.7 \\
\midrule
\multicolumn{13}{@{}l}{\textit{Closed-Source VLMs}} \\
Gemini-3.1-Flash    & 31.4 & 47.8 & 37.3 & \basebest{38.4} & 33.0 & 30.5 & 63.5 & 48.3 & \basesecond{21.8} & \basesecond{53.2} & \basebest{67.5} & 84.8 \\
GPT-4o              & 29.0 & 46.1 & \basesecond{40.0} & 32.0 & 26.1 & 32.1 & 68.4 & 41.1 & 20.4 & 49.5 & 44.5 & 76.1 \\
GPT-5.4-mini        & 30.9 & 49.9 & 38.1 & 32.1 & 46.2 & 33.0 & 67.0 & 56.1 & 18.9 & 51.6 & 45.5 & \basesecond{87.0} \\
Claude-Haiku-4.5    & 24.8 & 40.7 & 39.5 & 23.4 & 27.1 & 29.6 & 61.3 & 46.2 & \basebest{22.3} & 52.7 & 56.0 & 67.4 \\
Claude-Sonnet-4.6   & 31.9 & \basesecond{52.6} & 36.3 & 27.3 & \basebest{54.5} & 30.8 & 68.1 & 49.7 & 20.2 & 47.1 & 44.8 & 65.2 \\
\midrule
\multicolumn{13}{@{}l}{\textit{Post-trained VLMs}} \\
\textbf{ProcessData-SFT-Qwen} 
& \sftcell{\textbf{58.5}} & \sftcell{\textbf{82.7}} & \sftcell{\textbf{75.0}} & \sftcell{\textbf{92.5}} 
& \sftcell{\textbf{87.7}} & \sftcell{\textbf{45.4}} & \sftcell{\textbf{92.4}} & \sftcell{\textbf{63.4}} 
& \sftcell{\textbf{17.0}} & \sftcell{\textbf{51.0}} & \sftcell{\textbf{96.5}} & \sftcell{\textbf{97.8}} \\
\textbf{ProcessData-SFT-Intern} 
& \sftcell{\textbf{56.8}} & \sftcell{\textbf{81.9}} & \sftcell{\textbf{77.3}} & \sftcell{\textbf{92.8}} 
& \sftcell{\textbf{88.2}} & \sftcell{\textbf{45.3}} & \sftcell{\textbf{91.5}} & \sftcell{\textbf{66.3}} 
& \sftcell{\textbf{16.3}} & \sftcell{\textbf{55.4}} & \sftcell{\textbf{97.0}} & \sftcell{\textbf{95.7}} \\
\bottomrule
\end{tabular*}
\end{table*}

\FloatBarrier
\begin{takeawaybox}
Taken together, RoboProcessBench should be interpreted at the task-family level rather than through a single aggregate score. Model families capture different subsets of process cues, while primitive-local progress and temporal reasoning remain persistent weaknesses. This fragmented profile supports a decomposition into separable local-state, progress, temporal, outcome, and primitive-level subskills.
\end{takeawaybox}

\subsection{Effect of ProcessData-SFT Supervision}
\label{subsec:post_training}


ProcessData-SFT provides a broadly learnable supervision signal across model backbones. Both \textit{ProcessData-SFT-Qwen} and \textit{ProcessData-SFT-Intern} substantially improve over their corresponding zero-shot backbones and achieve strong performance on most local state, motion, and primitive-aware tasks. The gains are especially consistent on phase recognition (\textbf{\texttt{T1}}), contact detection (\textbf{\texttt{T2}}), bimanual coordination (\textbf{\texttt{T4}}), motion direction prediction (\textbf{\texttt{T3}}), motion-state recognition (\textbf{\texttt{T6}}), current primitive recognition (\textbf{\texttt{T10}}), next primitive prediction (\textbf{\texttt{T11}}), and primitive-chain restoration (\textbf{\texttt{T12}}). The fact that two different VLM backbones show similar improvement patterns suggests that the supervision is not tied to a single model family, but reflects visually learnable process cues constructed from physically grounded traces and primitive annotations. This is important for RoboProcessBench because it indicates that ProcessData-SFT is not merely an evaluation split with manually designed labels, but can also function as a training signal that transfers across different model architectures.

SFT improves both benchmark dimensions. Following the taxonomy of RoboProcessBench, we further summarize the effect of ProcessData-SFT under the two primary dimensions: \textit{static monitoring} and \textit{dynamic reasoning}. Compared with their corresponding base backbones, \textit{ProcessData-SFT-Qwen} and \textit{ProcessData-SFT-Intern} improve static monitoring by 42.7 and 44.4 points, respectively, reaching 77.2\% for both models. Dynamic reasoning also improves substantially, by 23.6 and 21.4 points, reaching 68.9\% and 69.5\%. This shows that ProcessData-SFT provides learnable supervision for both static interaction states and dynamic process cues. At the same time, the smaller gains on dynamic reasoning indicate that process evolution remains harder than recognizing local interaction states. This gap is expected because dynamic reasoning requires the model to infer motion, progress, temporal consistency, and primitive-level continuation from visual evidence, rather than only grounding visible interaction states in a single frame or local observation. Therefore, the static--dynamic comparison further supports the need for evaluating process-aware manipulation understanding at the task-family level instead of relying only on an aggregate score.

The improvement on primitive progress recognition (\textbf{\texttt{T5}}) is particularly important. \textit{ProcessData-SFT-Qwen} reaches \(45.4\%\) and \textit{ProcessData-SFT-Intern} reaches \(45.3\%\), improving by roughly 11 points over the strongest zero-shot baseline. This result suggests that local subprogress is not purely inaccessible to VLMs: structured process supervision can teach models to recognize some visual evidence of advancement within a manipulation primitive. At the same time, \textbf{\texttt{T5}} is far from saturated, indicating that progress estimation remains substantially harder than local state recognition.

The gains are not universal. Temporal ordering (\textbf{\texttt{T8}}) remains close to chance after post-training for both backbones, and temporal-priority prediction (\textbf{\texttt{T9}}) improves only for the InternVL-based model while remaining weak overall. Outcome prediction (\textbf{\texttt{T7}}) also improves less consistently than local state and primitive-aware tasks. This asymmetry clarifies what ProcessData-SFT currently teaches well: local process states, short-horizon motion cues, primitive transitions, and partial local progress. It does not yet solve robust temporal reconstruction or reliable outcome prediction. These remaining bottlenecks suggest that future VLM-based process evaluators may require stronger video modeling, explicit temporal objectives, memory over longer execution windows, or downstream progress-ranking validation.

Overall, these results show that ProcessData-SFT is a learnable substrate for process-aware understanding in robotic manipulation. It consistently improves two different VLM backbones on local state, motion, progress, and primitive-aware cues, indicating that many process labels derived from physically grounded traces correspond to visually learnable signals. At the same time, the remaining weakness on temporal ordering, temporal priority, and outcome prediction clarifies what current process supervision does not yet solve. This makes RoboProcessBench useful both for supervised adaptation and for diagnosing unresolved temporal and outcome-oriented reasoning bottlenecks.

\subsection{External reward evaluation on RoboRewardBench}
\label{subsec:external_reward}

We further study reward prediction on RoboRewardBench~\citep{lee2026roboreward}, an external benchmark of 2,831 robot episodes from OXE and RoboArena. These experiments evaluate reward prediction error.

\vspace{0.4cm}

\noindent\textbf{Cross-model association.} For 11 open-source baselines and two ProcessData-SFT models, we correlate the chance-normalized macro score over T5--T9 with negative group MAE. Positive correlation indicates lower reward error. Spearman correlation is $\rho$ = 0.818 for the 11 baselines and $\rho$ = 0.802 across all 13 models. This aggregate association suggests moderate casual relationships between process-aware understanding and reward labeling, indicating out framework's downstream utilities.

\vspace{0.4cm}

\noindent\textbf{Reward model adaptation.} Mixed training and replay calibration lower Group and OXE MAE for both RoboReward model scales. RoboArena changes are less consistent: 4B improves from 1.089 to 1.046, while 8B regresses from 0.996 to 1.018. 

\begin{table}[t]
\centering
\small
\setlength{\tabcolsep}{6pt}
\captionsetup{skip=10pt}
\caption{\textbf{External reward prediction.} RoboRewardBench MAE ($\downarrow$) before and after mixed training with ProcessData and replay calibration.}
\label{tab:external_reward_results}
\begin{tabular}{llccc}
\toprule
\textbf{Model} & \textbf{Group} & \textbf{OXE} & \textbf{RoboArena} \\
\midrule
RoboReward-4B & 0.845 & 0.847 & 1.089 \\
 w/ mixed + calibration & \textbf{0.457} & \textbf{0.431} & \textbf{1.046} \\
\midrule
RoboReward-8B & 0.665 & 0.768 & \textbf{0.996} \\
 w/ mixed + calibration & \textbf{0.426} & \textbf{0.399} & 1.018 \\
\bottomrule
\end{tabular}
\end{table}
\section{Limitations}
\label{sec:limitations}
\enlargethispage{4\baselineskip}

\begin{limitationbox}
Robo2Process depends on source-native signals to generate process-aware supervising signals at scale. Also, post-training gains on RoboProcessBench remain uneven across all tasks, especially for the temporal reasoning ones. We believe future work should lean towards these directions. 
\end{limitationbox}

\section{Conclusion}
\label{sec:conclusion}

We introduce Robo2Process, a trace-grounded framework that converts heterogeneous robotic execution data into process-aware supervision. Its instantiations, RoboProcessBench and ProcessData, cover four sources, 12 question families, 260 tasks, and 57,892 QA pairs. Evaluation of 16 VLMs reveals uneven process understanding. Post-training on two backbones improves local-state, motion, progress, and primitive-aware abilities, while temporal reasoning remains to be tackled. Out-of-distribution validation on RoboRewardBench provides further evidence that process scores associate with reward error and mixed training using ProcessData reduces it. We hope that Robo2Process will serve as a starting point to advance process-aware understanding in robotic systems.

\newpage

\bibliographystyle{unsrtnat}
\bibliography{references}

\end{document}